\documentclass[journal]{IEEEtran}
\usepackage{ulem}
\ifCLASSINFOpdf
   \usepackage[pdftex]{graphicx}
\else
   \usepackage[dvips]{graphicx}
\fi

\usepackage{amsmath}
\usepackage{amssymb}
\usepackage[vlined,ruled,linesnumbered]{algorithm2e}
\usepackage{subfig}
\usepackage{comment}
\usepackage{color}

\usepackage{gensymb}
\usepackage{multirow}
\usepackage{array}
\usepackage{xcolor}
\definecolor{Red}{rgb}{1,0,0}
\definecolor{Blue}{rgb}{0,0,1}
\definecolor{Green}{rgb}{0,0.5,0}
\definecolor{Gray}{gray}{0.4}

\newcommand{\todo}[1]{\textcolor{Red}{\textsf{\textbf{TODO:}~#1}}}

\newcommand{\oursystem}{LOPF}

\begin{document}
%
\title{Learning to Solve AC Optimal Power Flow by Differentiating through Holomorphic Embeddings}
%
%
%

\author{\IEEEauthorblockN{Henning Lange\IEEEauthorrefmark{1}, Bingqing Chen\IEEEauthorrefmark{2}, Mario Berg\'es\IEEEauthorrefmark{2}, Soummya Kar\IEEEauthorrefmark{2}}\\
\IEEEauthorblockA{\IEEEauthorrefmark{1} University of Washington, Seattle, WA, 98195, USA}\\
\IEEEauthorblockA{\IEEEauthorrefmark{2} Carnegie Mellon University, Pittsburgh, PA 15213, USA}
\thanks{}
\thanks{}
\thanks{}}

\markboth{LOPF: Learning Optimal Power Flow by Differentiating through Holomorphic Embeddings}%
{Shell \MakeLowercase{\textit{et al.}}: Bare Demo of IEEEtran.cls for IEEE Journals}
%



\maketitle



\begin{abstract}

Alternating current optimal power flow (AC-OPF) is one of the fundamental problems in power systems operation. AC-OPF is traditionally cast as a constrained optimization problem that seeks optimal generation set points whilst fulfilling a set of non-linear equality constraints -- the power flow equations. With increasing penetration of renewable generation, grid operators need to solve larger problems at shorter intervals. This motivates the research interest in learning OPF solutions with neural networks, which have fast inference time and is potentially scalable to large networks. The main difficulty in solving the AC-OPF problem lies in dealing with this equality constraint that has spurious roots, i.e. there are assignments of voltages that fulfill the power flow equations that however are not physically realizable. This property renders any method relying on projected-gradients brittle because these non-physical roots can act as attractors. In this paper, we show efficient strategies that circumvent this problem by differentiating through the operations of a power flow solver that embeds the power flow equations into a holomorphic function. The resulting learning-based approach is validated experimentally on a 200-bus system and we show that, after training, the learned agent produces optimized power flow solutions reliably and fast. Specifically, we report a 12x increase in speed and a 40\% increase in robustness compared to a traditional solver. To the best of our knowledge, this approach constitutes the first learning-based approach that successfully respects the full non-linear AC-OPF equations.
\end{abstract}


%

\section{Introduction}
The Optimal Power Flow (OPF) problem is fundamental to power systems operation \cite{fioretto2020predicting}. In general, OPF finds the optimal generation set points that minimize operation costs given a set of loads while at the same time satisfying physical and security constraints. The physicality of solutions is ensured by enforcing the power flow equations -- a set of non-linear equality constraints. 

The increasing penetration of distributed energy resources (DER) is posing new challenges for power system operation. Traditionally, generation schedules were updated at 5-min intervals \cite{fioretto2020predicting}. Due to the variability and uncertainty of renewable generation, system operators are required to update generation schedules more frequently. At the same time, system operators need to manage a growing number of smaller resources, resulting in a significantly larger problem than the conventional OPF problem \cite{guerrero2020towards}. Thus, the integration of DERs requires solving larger OPF problems at shorter time intervals. This motivates research on using function approximators, such as neural networks (NN), to learn solutions to the OPF problem, due to their fast inference and their potential to scale well to larger power networks \cite{owerko2020optimal}.

\begin{figure}
    \centering
    \includegraphics[width = \linewidth]{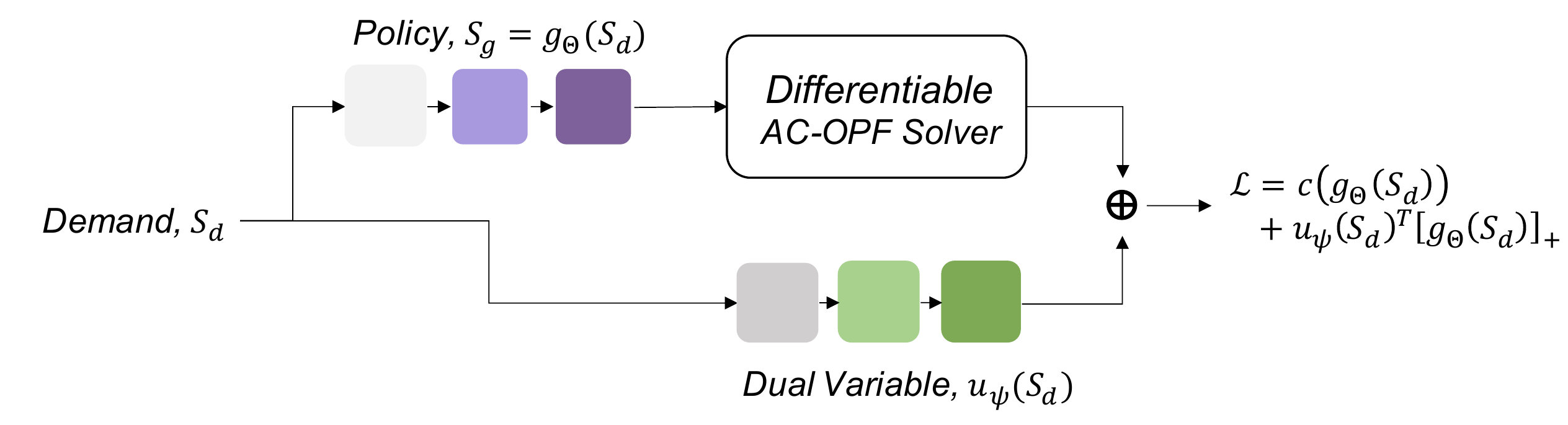}
    \caption{\textbf{Framework: } We propose a learning-based approach to AC-OPF problem by directly differentiating through a HELM solver. Thus, we learn a neural policy, $g_\Theta(S_d)$, that respects power flow equations. At the same time, We impose additional constraints, e.g., voltage magnitude and unit commitment, by learning dual variables corresponding to these constraints with another NN, $u_\Psi(S_d)$. Finally, we learn the policy end-to-end by optimizing the Lagrangian function, $\mathcal{L}$}
    \label{fig:framework}
\end{figure}

In order to deal with the non-linearity of the OPF problem, a common practice in power system operation is to linearize the grid dynamics, leading to the DC-OPF approximation. The DC-OPF problem is generally formulated as either a linear program (LP) or a quadratic program (QP), depending on the cost function.
Such a linearization can work reasonably well for small networks, but the resulting approximation error scales poorly with network size \cite{ferc2012history}. Furthermore, the assumptions required for linearizing the power flow equations are not valid for heavily-loaded or distribution networks \cite{chatzivasileiadis2018lecture} and the author of \cite{baker2019solutions} show that even under mild assumptions DC solutions are usually never AC feasible.
Ideally, the full non-linear AC power flow equations should be used. However, using the non-linear power flow equations comes with another challenge. Because the non-linear AC-OPF equality constraints have a fractal nature riddled with spurious roots, which act as attractors \cite{thorp1997load}, it is difficult to disambiguate physical from spurious non-physical solutions. As a result, convergence to the true solution cannot be guaranteed rendering existing solvers brittle.\\

In this paper, we propose a learning-based approach to the AC-OPF problem that alleviates the aforementioned issues by differentiating through a robust power flow solver, as shown in Figure \ref{fig:framework}. Our contributions can be summarized as follows. Firstly, we show that a power flow solver based on a holomorphic embedding of the power flow equations \cite{trias2012holomorphic} is differentiable in its arguments. The primary benefit of using Holomorphic Embedded  Load  Flow  method (HELM) is that it is much more likely to find the correct solution if it exists.  
Secondly, we can use this differentiable HELM solver as a computational layer in a NN. This allows us to learn a neural policy that solves the OPF problem, while respecting the power flow constraints. Furthermore, we show how this method can be extended to enforce convex security constraints, i.e., voltage magnitudes, as well as non-convex integer constraints, i.e., unit commitment. To validate our approach, we empirically show that it is faster and more robust (i.e., results in physically-viable solutions more often) on a 200-bus system in comparison to traditional solvers based on mixed-integer interior point methods (MATPOWER \cite{zimmerman2011matpower}). Specifically, our proposed method achieves a 12x increase in speed and a 40\% increase in robustness compared to MATPOWER.

\section{Related Work} 
\subsection{Challenges of Solving AC-OPF}

As described earlier, AC-OPF is traditionally posed as a constrained optimization problem, i.e. a cost function is minimized under network constraints oftentimes relying on some form of projected gradient descent \cite{carpentier1962contribution}.
Note that this problem formulation faces numerous computational difficulties. First, the non-linear equality constraint (\ref{eq:pf_const}) poses a challenge. In reality, (\ref{eq:pf_const}) is a necessary but not sufficient condition for the system to be in a physical state. There are assignments of nodal voltages $v$ that fulfill the power flow equations that. however, do not constitute a physical state \cite{tamura1983relationship,thorp1997load}. Optimization algorithms based on some form of projected gradient descent might be attracted to such a non-physical solution rendering them not robust. Advanced techniques like the Homotopy \cite{okumura1991solution} or Continuation \cite{milano2009continuous} methods alleviate but not fully remedy this problem, and can incur substantial computational costs. Furthermore, non-convex constraints create additional computational issues. Algorithms such as branch-and-bound that are typically employed to deal with integer constraints require solving multiple, and in the worst case exponentially many, relaxed linear programs~\cite{lawler1966branch}.\\
\subsection{Learning OPF}

Because larger OPF problems need to be solved more frequently, there is growing interest in learning to solve OPF problems with NNs. These approaches formulate OPF as the problem of learning a policy, which maps the demand configurations to generation set points. This policy could be learned from historical data, i.e., imitation learning, and/or by interacting with the system, i.e. reinforcement learning. 

The most common paradigm of existing solutions is to learn a mapping from demand assignments to optimal generation set points based on a data set provided by an external solver, such as MATPOWER \cite{zimmerman2011matpower}. This is analogous to behaviour cloning of an expert policy. In this case, the external OPF solver constitutes the expert policy \cite{chen2020learning}. Following \cite{gupta2020deep}, we refer to such a paradigm as \textit{OPF-then-learn}. Specifically, authors of \cite{guha2019machine, owerko2020optimal} pose AC-OPF as an end-to-end regression task. Specifically, in \cite{owerko2020optimal} graph neural networks (GNN) were adopted, which make predictions based on local information from neighboring nodes. Such an approach allows for a decomposition of the grid-level problem based on power system connectivity and, in principle, scales well to large networks. However, a significant limitation of such an approach is that the solutions may not adhere to physical and security constraints inherent to the physical system. Concretely, 49\% and 30\% of the solutions were infeasible when applied to the IEEE 30-bus and 118-bus test systems respectively \cite{guha2019machine}. 

This has motivated research on learning-based methods that respect constraints. The authors of \cite{chen2020learning, zhao2020deepopf+, singh2020learning} studied the problem in the context of DC-OPF. Since DC-OPF can either be posed as an LP or QP, the problem has amenable mathematical properties and permits for relatively simple solution strategies. More relevant to our work, \cite{zamzam2019learning, fioretto2020predicting} studied the problem in the context of AC-OPF. To ensure feasibility, in \cite{zamzam2019learning} generation set points produced by a NN were passed into a power flow solver. This speeds up the computation since the power flow equations are easier to solve than AC-OPF. In \cite{fioretto2020predicting}, the authors applied primal-dual updates to the Lagrangian dual function of the underlying constrained optimization problem, where a NN predicts the generation set points and the power system states. Notably, the approach proposed by \cite{fioretto2020predicting} produces more accurate and cost-effective solutions than those by DC-OPF approximation, which is commonly used in industry.

However, there are key limitations to the \textit{OPF-then-learn} paradigm. Firstly, a large labeled data set needs to be generated for training. Given the convergence issues of solving AC-OPF, some load configurations may not have a solution from commercial solvers. 
Secondly, the learned solution may not generalize to unseen scenarios and re-training the system on incoming demand assignments requires creating a new training set. Because of this, authors of \cite{gupta2020deep} proposed \textit{OPF-and-learn} as an alternative paradigm where generation cost is optimized directly by computing gradients through the OPF problem. While \textit{OPF-then-learn} is analogous to imitation learning, \textit{OPF-and-learn} can be seen as a one-step Markov Decision Process (MDP). However, the \textit{OPF-and-learn} approach proposed in \cite{gupta2020deep} relies on the DC-OPF linearization, which limits the accuracy of solutions and is not applicable to heavily-loaded or distribution networks. Furthermore, when assuming the DC-OPF linearization, creating a training set is trivial because optimal solutions can be obtained reliably and fast. Thus, by making the DC-OPF assumption, the circular dependence on training data is broken but a significantly easier problem is solved which limits the potential advantages of the \textit{OPF-and-learn} paradigm. 

To the best of our knowledge, this paper introduces the first \textit{OPF-and-learn} approach that respects the full non-linear power flow constraints as well as non-convex unit-decomittment and non-linear physical and security constraints.

\section{Proposed Learning Framework}
Let $c(v)$ be the cost associated with the assignment of nodal voltages $v \in \mathbb{C}^N$ and $N$ being the number of buses of the system. In the following, without loss of generality, we define the cost in terms of $v$ for presentational simplicity. However, numerous constraints need to be enforced, some of which are non-linear and non-convex, i.e. the load flow problem is traditionally posed as:

\begin{align}
\text{minimize w.r.t. $v$:\ }& c(v) \label{eq:lopf_erm_objective}\\
\text{subject to:\ }& S = S_{g} - S_{d} = diag(v)(Yv)^* \label{eq:pf_const}\\
 & k_i(v) \leq 0\\
 & h_i(v) = 0
\end{align}
with $S_d\in \mathbb{C}^N$ and $S_g\in \mathbb{C}^N$ being a demand and generation assignment, respectively, and $Y\in \mathbb{C}^{N \times N}$ being the bus admittance matrix.
For different demand assignments $S_d$, this optimization problem would be solved over and over again.\\

Our proposed learning-based formulation following the \textit{OPF-and-learn} paradigm is as follows: First, we note that the nodal voltages $v$ are a function of $S_d$ and $S_g$, i.e. $v = \mathfrak{v}(S_d,S_g)$ with $\mathfrak{v}$ solving the power flow equations (\ref{eq:pf_const}). Second, we introduce a function that is tasked with producing optimal generator configurations $S'_g$ as a function of the system state, in this case the demand $S_d$, in an expected risk minimization setting. Thus, $S_g' = g_\Theta(S_d)$ with $\Theta$ parameterizing the function $g$. Third, we assume knowledge of a data set $D$ containing historic demand assignments, i.e. a collection of possible states $S_d \in D$. Note that unlike \textit{OPF-then-Learn} approaches [cite,cite,cite] we do not assume knowledge of the corresponding optimal generator assignments. The goal is to estimate the parameters of the function $g$, in this case $\Theta$, that produce optimal generation assignments as a function of the demand. As an optimization objective and learning signal for the function $g$, we propose the following:
\begin{align}
\text{minimize w.r.t. $\Theta$:\ }& \sum_{S_d \in D} c(\mathfrak{v}(S_d,g_\Theta(S_d))) = \mathcal{L} \label{LOPF_obj}\\
\text{subject to:\ }& k_i(v) \leq 0 \label{const1}\\
 			    & h_i(v) = 0  \label{const2}
\end{align}
When compared to traditional optimization based approaches, this problem formulation has a number of advantages:
\begin{enumerate}
\item The non-linear power flow constraint (\ref{eq:pf_const}) vanishes. This increases robustness and avoids convergence to non-physical solutions given a differentiable and robust power flow solver.
\item As we will show later, optimization under non-convex constraints can be amortized, i.e. time training the system is spent once and after training, inference is extremely fast and solely requires a forward-pass through a NN even when respecting binary unit-commitment constraints.
\item Because training $g$ entails learning an optimal mapping between Euclidean spaces that represent demand and generation assignments respectively, the proposed approach exploits covariances between load flow problems and allows for the generalization to unseen problems.
\end{enumerate}

However, optimizing $\Theta$ w.r.t. (\ref{LOPF_obj}) poses challenges: If gradient descent is used for optimization, gradients need to be defined. Applying the chain rule to $\mathcal{L}$ yields: 
\begin{align*}
    \frac{\partial \mathcal{L}}{\partial \Theta} = \frac{\partial c}{\partial \mathfrak{v}} \frac{\partial \mathfrak{v}}{\partial g} \frac{\partial g}{\partial \Theta}
\end{align*}

Thus, the loss is differentiable if the cost function $c$, the voltage function $\mathfrak{v}$ and the actor function $g$ are differentiable. $g$ and $c$ can be usually assumed to be differentiable, however, the fact that computing the gradient through a power flow solver w.r.t. to the generation assignment $S_g$, i.e. $\frac{\partial \mathfrak{v}}{\partial S_g}$, is possible, might not be obvious. In section \ref{sec:helm}, we will show that computing the gradient through a robust power flow solver that represents the full non-linear AC-OPF equations, namely the Holomorphic Embedded Load Flow Method, is indeed possible. Doing this allows us to obtain the learning signal $\frac{\partial \mathcal{L}}{\partial \Theta}$.\\
Furthermore, the constraints i.e. (\ref{const1}) and (\ref{const2}) need to be enforced. In section \ref{sec:constraints}, we will show that an auxiliary function $\mathfrak{u}$ can be used to enforce the Karush-Kuhn-Tucker conditions ultimately allowing us to enforce arbitrary constraints similar but not identical to the approaches introduced in \cite{fioretto2020predicting, gupta2020deep}.\\
In section \ref{sec:bin_contraints}, we show how the proposed approach can be extended to handle non-convex, e.g. binary unit commitment constraints, by optimizing a variational lower bound whilst introducing little computational overhead during inference.\\
The approach is validated empirically on a 200 bus system. The experimental setup and results are described in section 5. In section 6, our findings are concluded and pathways for future work are laid out.

\section{Holomorphic Embedded Load Flow Method}
\label{sec:helm}
HELM was first proposed by Trias \cite{trias2012holomorphic,trias2015fundamentals} and was later extended in e.g. \cite{subramanian2013pv,wallace2016alternative}. HELM addresses the convergence issues of power flow solvers based on Householder's methods for root-finding such as the Newton-Raphson algorithm. Because the power flow equations have multiple roots but only a single physically realizable solution, these types of solvers are at risk to converge to a spurious, unstable or low voltage solution~\cite{deng2013convergence}. Whether or not the `correct' root is being found is usually dependent on the initial condition ~\cite{thorp1997load}. Because HELM does not require an initial guess or initial condition, it overcomes the ambiguity problem that Householder's solvers face, by performing analytical continuation from a known physically-realizable solution.
Analytical continuation also allows HELM to find the root that is on the same branch-cut as the previously known physically-realizable solution. This solution is unique because analytical continuation is unique when the function at hand is holomorphic. HELM treats the complex nodal voltages at each bus as holomorphic functions of a complex scalar $z$. These functions are evaluated at a point for which obtaining a physically-realizable solution is trivial (usually $z=0$) and are then continued to the solution at a desired point where the original power flow equations are recovered (usually $z = 1$).

Let $\mathcal{V}(z)$ be a function of the complex scalar $z$. Equation (\ref{eq:helm_embed}) then describes such a holomorphic embedding, i.e. obtaining a solution at $z = 0$ is trivial because no power is flowing and the original power flow equations (\ref{eq:pf_const}) are recovered at $z = 1$. See \cite{wallace2016alternative} for a proof that $\mathcal{V}(z)$ is indeed holomorphic.
\begin{equation}
Y\mathcal{V}(z) = \frac{zS^*}{\mathcal{V}^*(z^*)} \label{eq:helm_embed}
\end{equation}

In order to obtain the power series coefficients required for analytical continuation, $\mathcal{V}(z)$ and its reciprocal are approximated by a power series expansion, i.e.:
\begin{align}
\mathcal{V}(z) &= \sum_{n=0}^\infty c[n]z^n \label{eq:power_series}\\
\frac{1}{\mathcal{V}(z)} = \mathcal{W}(z) &= \sum_{n=0}^\infty d^*[n]z^n
\end{align}

Similar to traditional power flow solver such as Newton Raphson, in order to avoid overspecification of the problem, a slack bus is introduced: Let $Y^r \in \mathbb{C}^{N-1 \times N-1}$ be the reduced $Y$ matrix by removing the row and column of the slack bus and $y_s \in \mathbb{C}^{N-1}$ be the slack-row of $Y$ sans self-admittance. We assume that the voltage at the slack generator is $v_s + 0j$ with $v_s \in \mathbb{R}$.\\

For the $i$th row of (\ref{eq:helm_embed}) the following then holds:
\begin{align}
\sum_k Y^r_{ik} \sum_n^{\infty} c_k[n]z^n + (v_s+0j)y_s &= zS_i^*\sum_n^{\infty} d^*_i[n]z^n
\end{align}
\begin{align}
\text{Setting $z=0$:\ }&\sum_k Y^r_{ik} c_k[0]  = - v_s y_s \label{eq:z0}
\end{align}

Thus, solving the linear system in (\ref{eq:z0}) yields a solution at $z=0$. Note that this solution is physically-realizable because no power is flowing. Higher order power series coefficients can be obtained by equating coefficients of the same order and by making use of $(\sum_{n=0}^\infty c[n]z^n)  (\sum_{n=0}^\infty d^*[n]z^n) = 1$ which yields:
\begin{align}
d_k[0]  &= \frac{1}{c_k[0]} \label{eq:d0}\\
\sum_k Y^r_{ik} c_k[n]  &= S_i^*d_i[n-1] \label{eq:fi}\\
d_i[n] &= - \frac{\sum_{m=0}^{n-1} c_i[n-m]d_i[m]}{c_i[0]} \label{eq:di}
\end{align}
After obtaining power series coefficients, analytic continuation is performed to obtain a solution at $z=1$. However, since the radius of convergence is usually smaller than 1, analytical continuation is performed using Pad\'e approximants instead of directly evaluating (\ref{eq:power_series}). Pad\'e is a rational approximation of power series known to have the widest radius of convergence \cite{stahl1989convergence}. Analytical continuation by Pad\'e approximation is performed as follows: 
\begin{align}
\mathcal{V}_i(z) \approx R(z)={\frac  {\sum _{{j=0}}^{{m}}a_{i,j}z^{j}}{1+\sum _{{k=1}}^{{m}}b_{i,k}z^{k}}} \label{eq:pade_approx}
\end{align}
Approximants of order $m$, i.e. $a_i$ and $b_i$ can be obtained from the power series coefficients by solving a linear system of equations, specifically:
\begin{align*}
\begin{bmatrix}I & M(c_i)\end{bmatrix} \begin{bmatrix}a_i\\ b_i\end{bmatrix} = c_i
\end{align*}
with: $ M(c_i) = \\
\begin{bmatrix}0 & \hdots  & 0& 0 & 0& 0\\
 -c_i[1] & 0 & \hdots  & 0&  0& 0\\
 -c_i[2]& -c_i[1] & 0  &  \hdots  & 0& 0\\
 -c_i[2]& -c_i[2] & -c_i[1] & 0 & \hdots & 0\\
 \vdots & \vdots & \vdots & \vdots & \vdots & \vdots  \\
  -c_i[n]& -c_i[n-1] & -c_i[n-2] &  & \hdots & -c_i[m]\\
 \end{bmatrix}$ \\ 
 
 and $I$ being the identity matrix. Because we perform analytical continuation to $z=1$, plugging the obtained coefficients into (\ref{eq:pade_approx}) yields: $V_i \approx \sum_{j=0}^m a_{j,i} / (1+\sum_{j=0}^m b_{j,i})$.

\subsection{Differentiating through HELM}
In the following we will view HELM as a function that maps complex nodal power to complex nodal voltages, i.e. $v = \mathfrak{v}(S_d,g_\Theta(S_d))$. We will show that $\mathfrak{v}$ is not holomorphic but $\mathbb{R}$-differentiable in $\Theta$ ultimately allowing us to compute gradients w.r.t. the parameters of the actor function $g$. By making use of the chain-tule, the strategy is to decompose HELM into a succession of functions and show that each function is $\mathbb{R}$-differentiable. Specifically, we decompose HELM into its algorithmic steps, i.e. $\mathfrak{v}(S_d,g_\Theta(S_d)) = f_v \circ f_{ab} \circ f_{c,n} (S_d-g_\Theta(S_d))$ with $f_{c,n}$ computing power series coefficients, $f_{ab}$ computing Pad\'e approximants and $f_v$ computing voltage phasors given Pad\'e approximants. We then show that $f_{ab}$,$f_{c,n}$ and $f_{v}$ are $\mathbb{R}$-differentiable. Note that $f_{c,n}$ is a recursive function and that writing its gradient out would be tedious. But its gradients can be computed efficiently using the backpropagation algorithm and the implementation is trivial in any deep learning frameworks with automatic differentiation capability, e.g. \texttt{PyTorch} and \texttt{TensorFlow}.

As stated earlier, HELM first computes the power series coefficients followed by Pad\'e approximation. The power series coefficients $c[n]$ and $d[n]$ are obtained in alternating fashion: Let $f_{c,n}$ and $f_{d,n}$ be the function that produces $c[n]$ and $d[n]$ respectively. Note that $f_{c,n}$ requires knowledge of the previous $d$-coefficient and $g_\Theta$, whereas $f_d$ is a function of all previous $c$- and $d$-coefficients:
\begin{align}
\text{corrs. to (\ref{eq:fi}): } & f_{c,n}(x)  = (Y^r)^{-1}f_{d,n-1}(x)x^* \label{eq:corr14}\\
\text{corrs. to (\ref{eq:di}): } & f_{d,n}(x)  = \frac{\sum_{m=0}^{n-1} f_{c,n-m}(x)f_{d,m}(x)}{f_{c,0}(x)} \label{eq:corr15}
\end{align}
Because of the complex conjugation in (\ref{eq:corr14}), $\mathfrak{v}$ is not holomorphic in $x$, and $\Theta$, if $x = S_d-g_\Theta(S_d)$. However, it is easy to see that, by induction, (\ref{eq:corr14}) and (\ref{eq:corr15}) are $\mathbb{R}$-differentiable when $f_{c,0}$ and $f_{d,0}$ are $\mathbb{R}$-differentiable which is easy to see from (\ref{eq:z0}) and (\ref{eq:d0}).\\
After obtaining the power series coefficients, Pad\'e approximants $a$ and $b$ are calculated. Note that this also only includes solving a linear system of equations, i.e. \begin{align*}
f_{ab}(x) = \begin{bmatrix}a\\ b\end{bmatrix} = \begin{bmatrix}I & M(x)\end{bmatrix}^{-1}  x
\end{align*} which is differentiable. Then $f_v$ includes only a summation and fraction, i.e:
\begin{align*}
f_v(\begin{bmatrix}a\\ b\end{bmatrix}) = \sum_{i=0}^m a_i / (1+\sum_{i=0}^m b_i)
\end{align*}
Although, $\mathfrak{v}(x) = f_v(f_{ab}(f_{c,n}(x)))$ is not holomorphic, it is $\mathbb{R}$-differentiable in its argument $x$ and, when applied to $x = S_d-g_\Theta(S_d)$, it is $\mathbb{R}$-differentiable in $\Theta$.

\section{Enforcing Constraints}
\label{sec:constraints}
\subsection{A priori constraints}
As stated earlier, we treat the generation assignment $S_g$ as the output of a parameterized function $g_\Theta$. Because of the reasoning laid out earlier, we require $g$ to be differentiable and because of recent successes of NNs in non-linear optimization, we choose $g$ to be a NN with a penultimate sigmoidal layer \cite{zamzam2019learning}. We incorporate the generation limits of the generators into the output layer of the NN and therefore enforce generation limits by construction. Let $\sigma \in (0,1)^{2N_g}$ be the penultimate layer with $N_g$ being the number of generator buses. Thus, every generator is associated with two neurons, i.e.:
\begin{align*}
g_\Theta(S_d)_i = (S_g)_i &= (P_i^{max} - P_i^{min})\sigma_i + P_i^{min}\\
 &+ j(Q_i^{max} - Q_i^{min})\sigma_{i+N_g}  + jQ_i^{min}
\end{align*}
with $P^{max}_i$,$P^{min}_i$,$Q_i^{max}$ and $Q_i^{min}$ being the active and reactive generation limit respectively. Because $\sigma$ is bounded by $(0,1)$ non-slack generation limits cannot be violated. However, other constraints such as e.g. voltage magnitude or thermal line limits are not enforced by construction. That is why, in the next section we show how to enforce what we call \emph{a posteriori} constraints, i.e. constraints whose violation is only known after evaluating $\mathfrak{v}$.

\subsection{A posteriori constraints}
We adapt ideas from mathematical optimization to enforce arbitrary constraints on $v$. In mathematical optimization, the Karush-Kuhn-Tucker conditions (KKT-conditions) are necessary conditions for a solution to be optimal~\cite{gordon2012karush}. Given the optimization problem (\ref{eq:lopf_erm_objective}) expressed in terms of $v$, the KKT conditions state that a solution $v'$ is locally optimal under some regularity conditions when there exist $\mu_i$ such that:
\begin{itemize}
\item $\forall_i \mu_i \geq 0$ (Dual feasibility)
\item $\forall_i \mu_i k_i(v') = 0$ (Complementary slackness)
\item $\forall_i k_i(v') \leq 0$ (Primal feasibility)
\item $0 =\nabla f(v') + \sum_i \mu_i \nabla k_i(v')$ (Stationarity)
\end{itemize}
Note that, without loss of generality (because any equality constraint can be expressed as two inequality constraints) and for notational convenience, we restrict the optimization problem to only have inequality constraints.

 However, as stated earlier, we are not interested in the solution of a single constraint optimization problem but instead in solutions to all instances of a class of optimization problem. In this case, $S_d$, i.e. the demand assignment, specifies the instance of the optimization problem whereas the network topology, i.e. admittance matrix $Y$, specifies the class. First, we note that the KKT-multipliers are dependent on the instance of the optimization problem, thus instead of introducing a scalar $\mu_i$, we introduce a scalar-valued function $u_\psi(S_d)$. In order to enforce dual feasibility by construction, we choose $u$ to be a NN with soft-plus output parameterized by $\psi$.  Furthermore, let $\mathfrak{g}_\Theta^{S_d} = \mathfrak{v}(S_d, g_\Theta(S_d))$, $(\mathfrak{u}_\psi^{S_d})_i = u_\psi(S_d)_i$ the $i$th output of $u$ and $k_i^+(v) = \max(k_i(v),0)$. We will now introduce a learning criterion and show that local optima of this criterion fulfill the KKT-conditions for instances of the class contained in the training set. As a learning criterion we propose:
\begin{align}
\mathcal{L}(S_d) = c(\mathfrak{g}_\Theta^{S_d}) + \sum_i (\mathfrak{u}_\psi^{S_d})_i k^+_i(\mathfrak{g}_\Theta^{S_d}) \label{minimal_loss}\\
\forall_{S_d \in D}\max_\psi\{ \min_\Theta\{ \mathcal{L}(S_d) \} \} \label{eq:kkt_objective}
\end{align}
We will now show that, after convergence, for all $S_d \in D$, $v' = \mathfrak{g}_\Theta^{S_d}$ is locally optimal under some regularity constraints, i.e. it fulfills the KKT-conditions and furthermore, that the KKT-multipliers for which the KKT-conditions hold are:
\begin{align}
\mu_i = \begin{cases} (\mathfrak{u}_\psi^{S_d})_i &\text{if\ } k_i(\mathfrak{g}_\Theta^{S_d}) = 0 \label{eq:lag_mult}\\
0 &\text{else}
\end{cases}
\end{align}

\begin{itemize}
\item Dual feasibility: $\mu_i$ is dual feasible by construction: it is either $0$ or greater than $0$ because it is the output of a soft-plus NN.
\item Complementary slackness: Follows directly from (\ref{eq:lag_mult})
\item Primal Feasibility: Since (\ref{eq:kkt_objective}) converged, we know that $\frac{\partial L}{\partial (\mathfrak{u}_\psi^{S_d})_i} = 0$ and since $\frac{\partial L}{\partial (\mathfrak{u}_\psi^{S_d})_i} = k^+_i(v') = 0$, $v'$ must be primal feasible. Or in other words: if $v'$ was not primal feasible, $k^+_i(v') > 0$ but then the maximization step of (\ref{eq:kkt_objective}) could have increased $L$ by increasing $\mu_i$ which is a contradiction to the assumption that (\ref{eq:kkt_objective}) has converged.
\item Stationarity: Follows directly from the assumption that (\ref{eq:kkt_objective}) has converged. Note that substituting $k^+_i$ for $k_i$ does not have an influence because if $k_i(v) \neq 0$ then the corresponding $\mu_i = 0$ (complementary slackness) and when $k_i(v) = 0$ then $\nabla k_i(v) = \nabla k_i^+(v)$
\end{itemize}

Note that the detour of substituting $k_i(v)$ for $k_i^+(v)$ improves the performance substantially. Without the substitution, because more constraints are complied with initially, the NN drives the outputs before the soft-plus non-linearity to $-\infty$ in order to make the corresponding $\mu_i$ equal to 0. The output units are then `dead' and, because the gradient of the output non-linearity is close to 0, will always stay 0.

\subsection{Enforcing Physicality}
\label{sec:enforcing_phys}
So far, we have shown how to enforce `a priori'-constraints, i.e. constraints whose violation is known before inferring nodal voltages, by construction, as well as `a posteriori'-constraints, i.e. constraints whose violation is known after inferring nodal voltages, by introducing a learning objective that, after convergence, will enforce the KKT-conditions. However, we have not yet shown how to keep the function $g$ in the physical regime, i.e. prevent $g$ from producing a generation assignment $S_g$ for some $S_d$ such that there is no $v$ that fulfills the power flow equations (\ref{eq:pf_const}). An extreme example of a non-physical tuple $(S_d, S_g)$, for any demand assignment $S_d$ for which $\sum_i real(S_d)_i > 0$ is $S_g = \vec{0}$. 

First, we note that HELM will always produce complex nodal voltages even for non-physical tuples. However, for non-physical tuples the power flow equations (\ref{eq:pf_const}) will not hold, i.e. there is a mismatch between the RHS and LHS of (\ref{eq:pf_const}). We quantify this mismatch by defining: 
\begin{align*}
\epsilon(v) = ||S_{g} - S_{d} - \text{diag}(v)(Yv)^*||_\infty
\end{align*} 

\begin{figure}
\includegraphics[width=\linewidth]{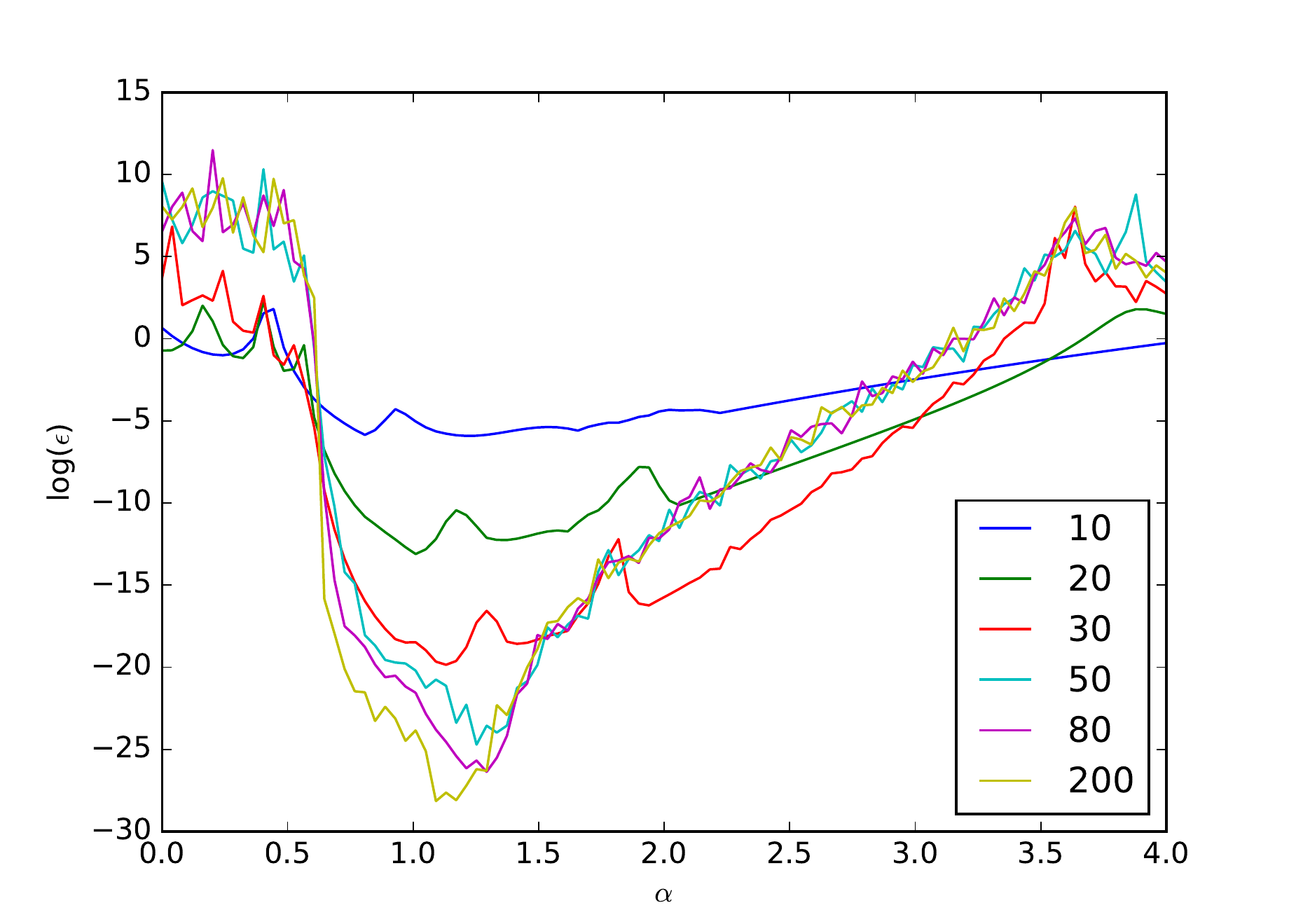}
\caption[\oursystem~: The $log$-mismatch $\epsilon$ as a function of $\alpha$]{The $log$-mismatch $\epsilon$ as a function of $\alpha$, i.e. $\epsilon(\mathfrak{v}(S_d,\alpha S_g))$. $\alpha$ scales a physical solution, i.e. when $\alpha = 1$ the corresponding $\epsilon$ is small. The number of HELM iterations $n$ is color coded.}
\label{fig:errors}
\end{figure}

The goal now is to enforce that $\epsilon(v) < \xi$ with $\xi$ being some parameter which specifies when a power flow solution is deemed physical. Note that because $\epsilon$ is a function of $v$, in principle, an additional inequality constraint could be introduced, i.e. $k_i(v) = \epsilon(v) - \xi \leq 0$ and one could try to enforce this constraint as an \emph{a posteriori} constraint as described earlier. However, in our experience this approach struggles, i.e. the learning objective usually does not converge. Figure \ref{fig:errors} gives an intuition as to why this is the case. Figure \ref{fig:errors} shows $\log(\epsilon)$ as a function of $\alpha$ on a 200 bus system. $\alpha$ scales the generation $S_g$ of a physical tuple $(S_d,S_g)$, i.e. the y-axis shows $\log(\epsilon(\mathfrak{v}(S_d,\alpha S_g)))$. Note that when $\alpha$ is either small or big ($<0.5$ or $>3.5$), $\epsilon$ is close to flat and therefore the gradient of $\epsilon$ is close to 0. After randomly initializing the the parameters of the function $g$, its guesses about optimal generation assignments will naturally be bad which corresponds to scaling the optimal generation assignment with a small or big $\alpha$. However, the function cannot improve its guesses by gradient descent because the gradient will be close to 0.

\begin{figure}
\includegraphics[width=\linewidth]{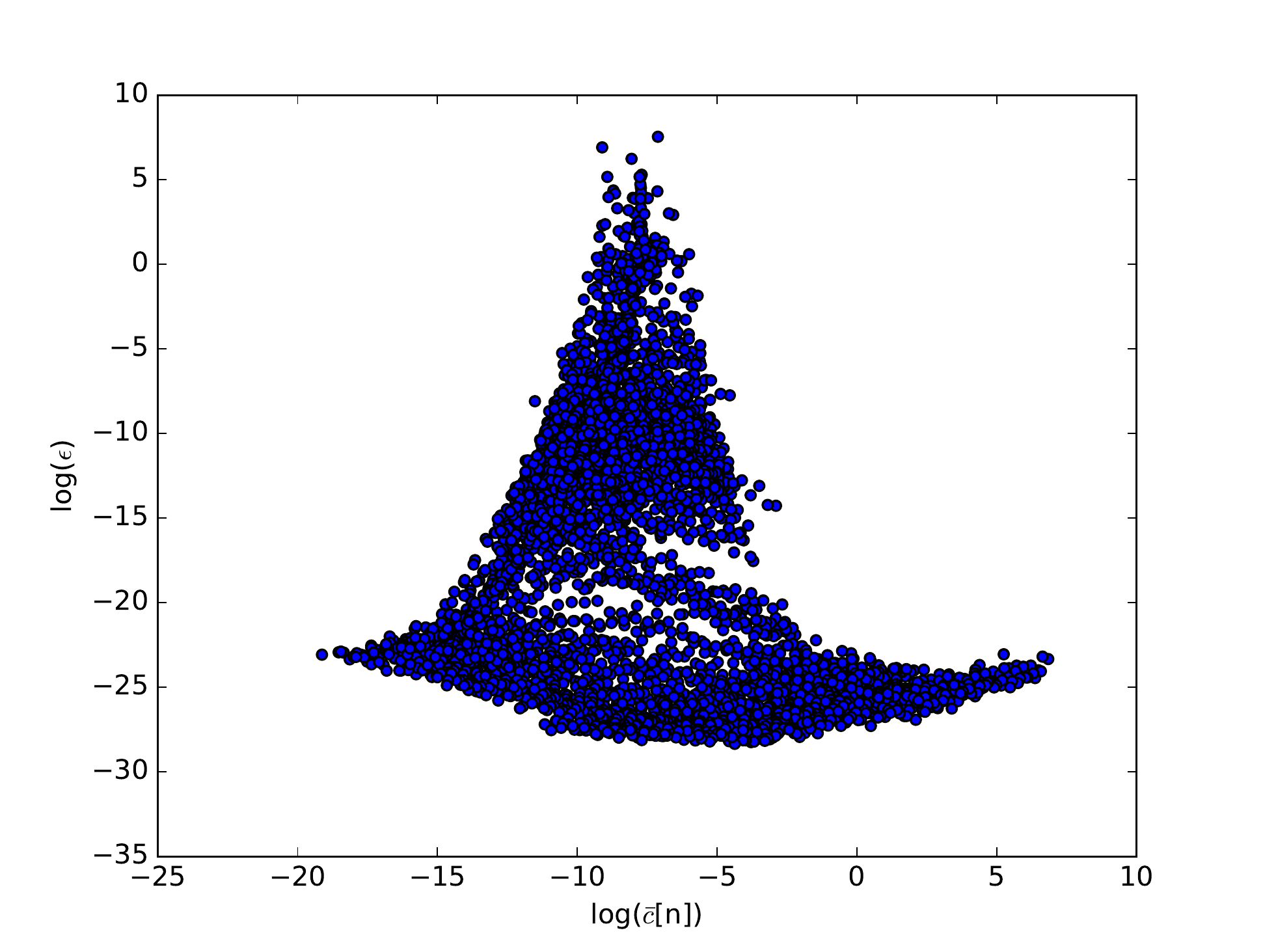}
\caption[\oursystem: Small power series coeffients imply small error.]{Small $\log \bar{c}[n]$ is a sufficient condition for small $\log(\epsilon)$. Specifically, when $\log \bar{c}[n] > -15$ then $-20 < \log(\epsilon) < -25$. However, the opposite is not true, i.e. small $\log(\epsilon)$ does not bound $\log \bar{c}[n]$.}
\label{fig:ceps}
\end{figure}

In order to overcome this problem, we propose to optimize a proxy of the actual mismatch function $\epsilon$. Note that an indicator of whether or not a solution is physical is whether or not the power series coefficients $c_i[n]$ have converged to 0. Let $\bar{c}[n]$ be the mean $n$th power series coefficient of all voltages, i.e. $\bar{c}[n] = \sum_i c_i[n]/N$. Figure \ref{fig:ceps} shows a scatter-plot of $\log \bar{c}[n]$ and $\log \epsilon$. Empirically, one can see that small $\bar{c}[n]$ is a sufficient condition for small $\epsilon$, however not a necessary condition. That is, a small $\bar{c}[n]$ implies small $\epsilon$ but not vice versa. Thus in order to enforce physicality, $\bar{c}[n]$ can be minimized as a proxy for $ \epsilon$. However, one might think that optimizing $\log \bar{c}[n]$ is unnecessarily restrictive, i.e. it excludes solutions where the power series coefficients did not converge to 0 but the corresponding $v$ nevertheless fulfill the power flow equations. But, as we will show later, imposing voltage magnitude constraints naturally enforces physicality and additionally minimizing $\log \bar{c}[n]$ is only required after the actor function $g$ was first initialized in order to `nudge' the it into the physical regime.

\section{Binary Constraints}
\label{sec:bin_contraints}
Binary constraints naturally occur in optimal power flow when incorporating the possibility of completely shutting down generators. Introducing this constraint also known as unit commitment makes generation limit constraints non-convex, i.e. $0$ becomes a possible generation assignment, even though points between $0$ and $P^{min}$ are not valid. Typically, optimal power flow solvers employ mixed integer programming techniques such as \emph{branch and bound} or \emph{branch and cut}~\cite{lawler1966branch} algorithms to tackle this problem. However, these algorithms can incur substantial computational cost, i.e. every branch requires solving an LP relaxed optimal power flow problem and there are exponentially-many branches in a worst-case scenario.

However, using the problem formulation introduced here, because the constraint is \emph{a priori} and can be enforced by construction, we can reduce the non-convex constraint into the problem of inferring the mode of a probability distribution $P$ over binary configurations. As we will show later, because inference in $P$ is intractable, we optimize a variational bound, i.e. we introduce a variational distribution $Q_\phi$ for which posterior inference is tractable and choose the variational parameters $\phi$ in such a way that $Q_\phi$ best approximates $P$. Specifically, we built on recent advances in Bayesian inference, specifically Variational Inference~\cite{wainwright2008graphical} and train a variational distribution $Q_\phi$ parameterized by a NN. See \cite{blei2017variational,zhang2017advances} for recent reviews of Variational Inference.

We begin by showing that computing the optimal binary configuration is equivalent to computing the mode of a distribution $P$. Let $b\in\{0,1\}^{N_g}$ be the vector describing which generators are turned \emph{on} or \emph{off} and $p(b,S_d)$ be an exponential distribution, $p(b|S_d)$ its posterior (Boltzmann distribution) and $L$ be the loss as defined in (\ref{minimal_loss}), i.e. 
\begin{align}
p(b, S_d) &= \lambda \exp{- \lambda L(b\cdot S_d)}\\
p(b|S_d) &= \frac {\exp{- \lambda L(b\cdot S_d)}}{\sum _{b' \in B}{\exp{- \lambda L(b'\cdot S_d)}}} \label{p_dist}
\end{align}
It is easy to see that computing the mode of (\ref{p_dist}), i.e. $\arg\max_b p(b|S_d)$ is equivalent to choosing the binary configuration that results in the smallest loss. However, na\"ive evaluation of the mode is usually intractable, because of the intractable denominator. Na\"ively computing the mode of (\ref{p_dist}) is equivalent to brute-force search, i.e. enumerating all possible latent configurations and picking the one with the smallest error. However, in the following, we show how ideas from Variational Inference~\cite{wainwright2008graphical} can reduce the computational burden of inference.

We introduce a variational distribution $Q_\phi$ whose posterior is tractable. Specifically, we choose $q(b|S_d)$ to be a multi-variate Bernoulli distribution and ensure tractability with ideas introduced in \cite{lange2018factornet}. Note that $Q_\phi$ is parameterized with a NN, therefore ensuring that inference at test-time is fast. As a learning signal for the parameters of the auxiliary posterior distribution $\phi$, we choose the Evidence Lower Bound defined by:

\begin{align}
L_{BO}(\phi) &= \mathbb{E}_{q_\phi(b|S_d)} \log \frac{p(b,S_d)}{q_\phi(b|S_d)} \label{elbo}\\
&= \log p(S_d) - D_{KL}(q_\phi(b|S_d) || p(b|S_d))
\end{align}

Note that optimizing (\ref{elbo}) does not require knowledge of the intractable posterior of $P$ but nevertheless allows for minimizing a divergence measure between the true ($P$) and auxiliary posterior ($Q$). Thus, after training, in order to obtain an approximation of the mode of $P$, because $P$ and $Q$ will be maximally similar, posterior inference is performed on $Q$ instead.
However, the price for this `trick' is increased variance. It can be shown that the stochastic gradient estimator of (\ref{elbo}) w.r.t. $\phi$ is an unbiased but higher variance estimator of the KL-divergence~\cite{mnih2014neural}. In order to combat variance, a decades-old variance reduction technique is employed, namely sampling without replacement. Sampling without replacement from $Q$ is not trivial. However, there is a considerable body of preexisting work that we make us of. The sampling scheme introduced in \cite{shah2018without} with slight modifications is employed. Specifically, instead of using the Pareto sampler as the underlying sampling mechanism, a slightly slower but more accurate elimination sampler introduced in \cite{deville1998unequal} is used.

\begin{figure}
\includegraphics[width=0.95\linewidth]{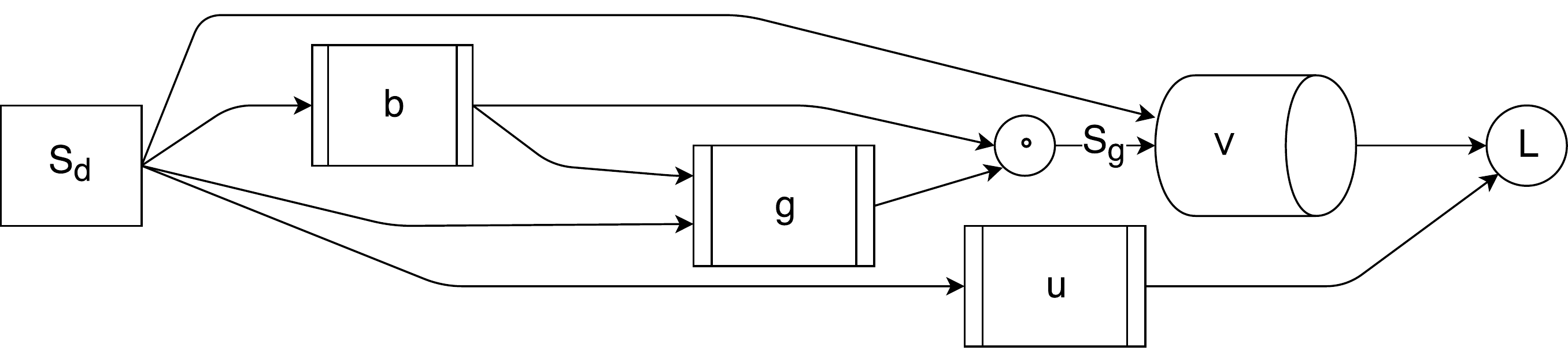}
\caption[\oursystem: Graphical depiction of the algorithmic pipeline]{A graphical depiction of the \emph{\oursystem}-pipeline. NNs $b$ and $g$ are fed the complex demand $S_d$ and tasked with producing the optimal binary activation vector and generator configuration respectively. Because the voltages are a function of demand and generation, both are fed into the HELM based power flow solver $v$. The loss $L$ is computed based on the resulting voltages. In order to ensure that control and network inequality constraints are satisfied, a third NN is tasked with predicting Lagrange multipliers. Because HELM is differentiable, the whole pipeline can be optimized jointly.}
\label{fig:pipe}
\end{figure}

In order to obtain an approximation of the mode of the true posterior, because $Q$ allows for drawing samples efficiently, $S$-many samples are drawn from $Q$. Then, in order to approximate the mode of $P$, out of the $S$-many binary configurations sampled from $Q$, the one which results in the smallest generation cost is chosen. Note that the optimal configuration of generators is dependent on the binary configuration, thus $b$ should additionally be fed into the actor function $g$. Figure \ref{fig:pipe} shows a graphical depiction of the proposed data pipeline.

\section{The \emph{\oursystem}-algorithm}
\label{sec:LOPFalgo}
In this section we summarize the resulting algorithm, we call \emph{\textbf{L}earning \textbf{O}ptimal \textbf{P}ower \textbf{F}low}, or short \emph{\oursystem}. The algorithm iterates over batches of the data set $D$ making updates to the three constituent NNs $g$, $b$ and $u$. It is described in pseudo-code in Algorithm \ref{algo:lopf}. For notational convenience, we define a function $solve$:
\begin{align*}
S_g &= b \cdot g_\Theta(S_d, b)\\
solve(S_d,b)
&=
\begin{pmatrix}
\epsilon(\mathfrak{v}(S_d,S_g)) \\
 c(\mathfrak{v}(S_d,S_g)) + \sum_i (u_\psi(S_d))_i k^+_i(\mathfrak{v}(S_d,S_g)) \\
N^{-1}\sum_i (f_{c,n}(S_d - S_g))_i
\end{pmatrix}^T
\end{align*}

\begin{algorithm}[htb]
\SetKwInOut{Input}{input}\SetKwInOut{Output}{output}
\Input{data set $D$ }
\Output{Trained model parameters $\Theta, \phi$ and $\psi$}
Initialize $\Theta, \phi$ and $\psi$ randomly\;
\While{not converged}{
	\For{number of subsets of $D$}{
		select $d \subset D$\;
		\For{$S_d \in d$}{
			$\overline{G} = \{\}; G = \{\};$ \;
			$B \sim q(b|S_d)$ (without replacement)\;
			\For{$b' \in B$}{
				$\epsilon, L, c \leftarrow solve(S_d,b')$\;
				\lIf{$\epsilon < \xi$}{$G \leftarrow G \cup \{L\}$}
				\lElse{$\overline{G} \leftarrow \overline{G} \cup \{c\}$}
			}
			Compute $L_{BO}$ based on (\ref{elbo})\;
			
		}
		Maximize $L_{BO}$ w.r.t. $\phi$\;
		Maximize $\sum_{L \in G} L$ w.r.t. $\psi$\;
		Minimize $\sum_{L \in G} L$ w.r.t. $\Theta$\;
		Minimize $\sum_{c \in \overline{G}} c$ w.r.t. $\Theta$\;
	}
}
\caption{\emph{\oursystem}-Algorithm in pseudo-code}
\label{algo:lopf}
\end{algorithm}

\begin{figure}[ht]
    \centering
    \includegraphics[width=\linewidth]{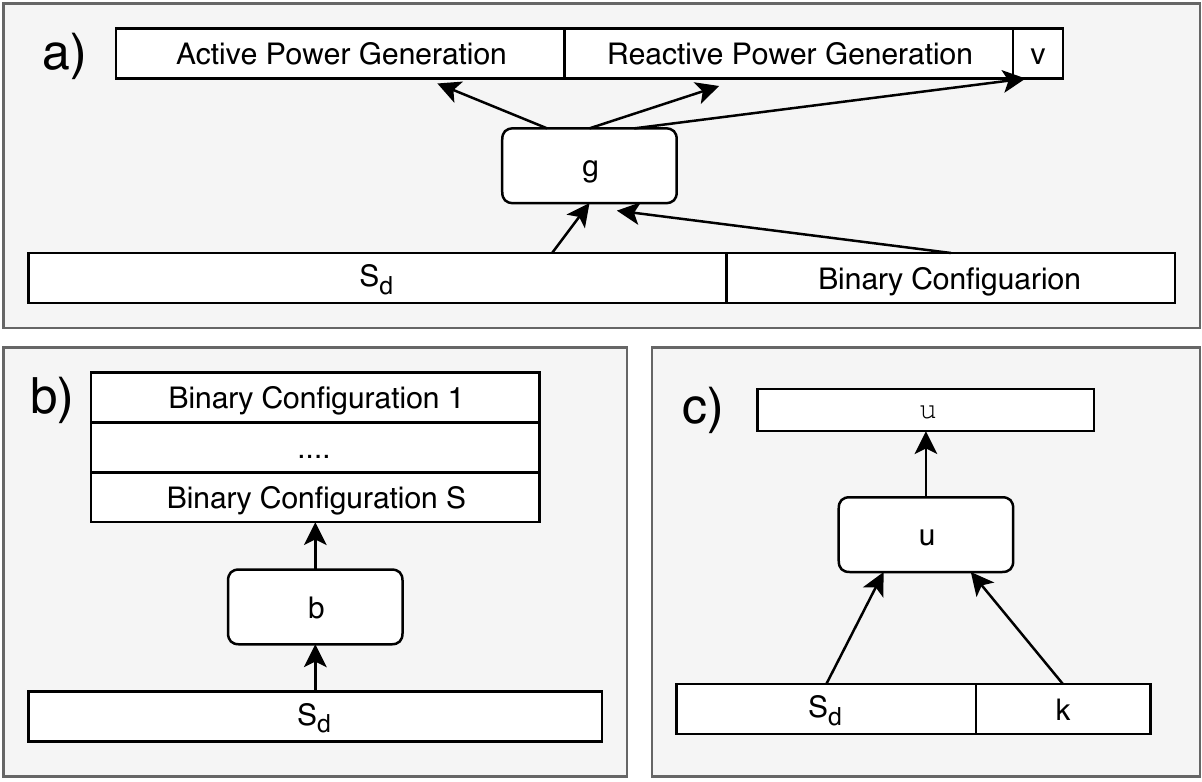}
    \caption[\oursystem: Graphical depiction of the constituent NNs.]{The input and output relationships of the three constituent NNs. a) The NN produces active and reactive power generation for non-slack generators as well as the voltage at the slack bus given the demand $S_d$ and binary configuration produced by the $b$-network. b) The binary-network that parameterizes an auxiliary distribution. Note that the network produces multiple binary configuration by sampling from the auxiliary distribution. c) The Lagrange-network that produces a proxy of the Lagrange multipliers. Note that the constraint-violation magnitude $k$ is additionally fed into the network to ease learning.}
    \label{lopf:constituent_networks}
\end{figure}

Figure \ref{lopf:constituent_networks} shows a graphical depiction of the input/output relationships of the individual networks. Note that the network $g$ not only produces active and reactive generation assignments for non-slack generators but also the voltage at the slack bus. Furthermore, the magnitude by which constraints are violated, denoted by $k$, are fed into the network $u$ that produces a proxy of the Lagrange multipliers. Additionally feeding $k$ into the $u$-network eases and speeds up learning considerably.

\section{Experiments}
\label{sec:LOPFexperiments}
Since this work introduces a learning based approach to the problem of ACOPF, the performance of the algorithm is evaluated similar to how the performance of reinforcement learning agents is evaluated, i.e. an empirical evaluation strategy is employed. Specifically, given a held out test set of load flow problems that the system was not presented with during training, the generation cost and the result of whether or not the system was able to find a feasible solution are recorded.\\

The requirements for feasibility excluding those that are met by construction are the following:
\begin{itemize}
    \item Log-mismatch between the RHS and LHS of the power flow equations (\ref{eq:pf_const}), i.e. $\epsilon$, must be smaller than $-10$.
    \item Slack active and reactive generation are within limits
    \item Non-slack voltage magnitude constraints are met
\end{itemize}

The experiments were conducted on the 200 bus Illinois IEEE test case~\cite{zimmerman2011matpower}. 
However, since the IEEE test cases only contain a single demand assignment, the demand base case was superimposed by temporal patterns extracted from the RE Europe data set \cite{re_europe}. 
RE Europe data set contains historical demand for 3 years at an hourly interval. Let $S_{d'} \in \mathbb{C}^{200}$ be the base demand taken from test case and $x_t \in \mathbb{R}^{200 \times 26280}$ be the temporal demand patterns taken from RE Europe data set. The temporal patterns were imposed such that the mean demand of every node is equal to the demand in the test case and such that the ratio between mean and standard deviation as seen in the RE Europe data set is preserved. The data set was separated into training (20.280 data points) and test set (6000) when conducting experiments.\\
The NNs used in this experiments constitute standard fully connected three-layer networks with intermediate $tanh$ activations. All intermediate layers have $512$ hidden units.

\begin{figure*}[!ht]
    \centering
    \includegraphics[width=0.85\linewidth]{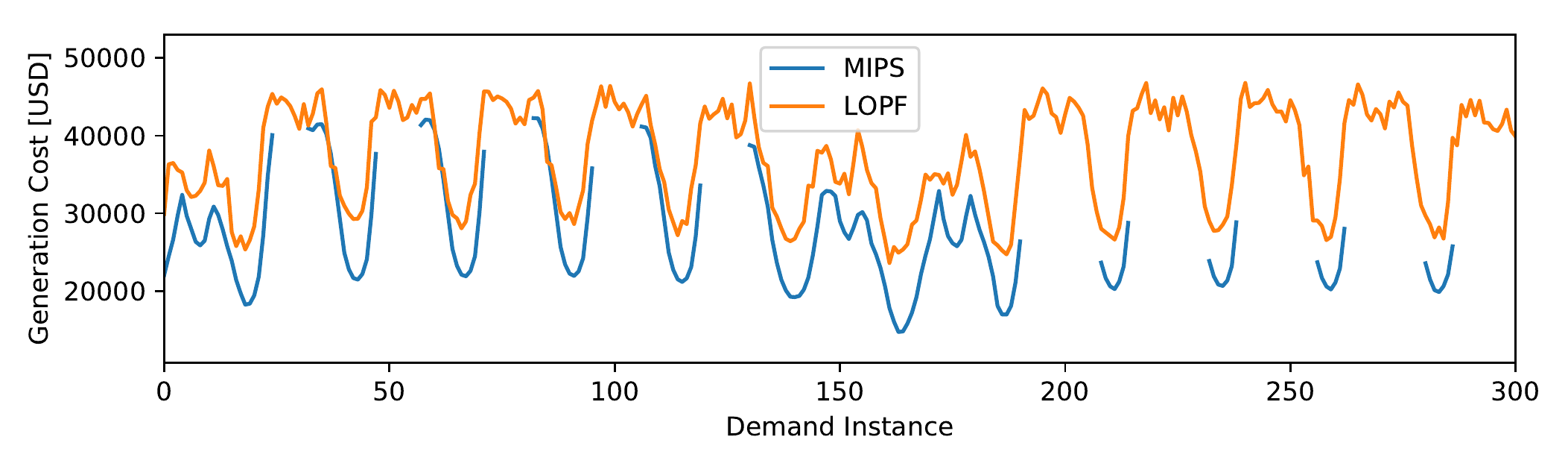}
    \caption[\oursystem: Comparison of generation cost.]{Comparison of generation cost on the first 300 load flow problems of the test set.}
    \label{lopf:cost}
\end{figure*}

\begin{figure*}[!ht]
    \centering
    \includegraphics[width=0.48\linewidth]{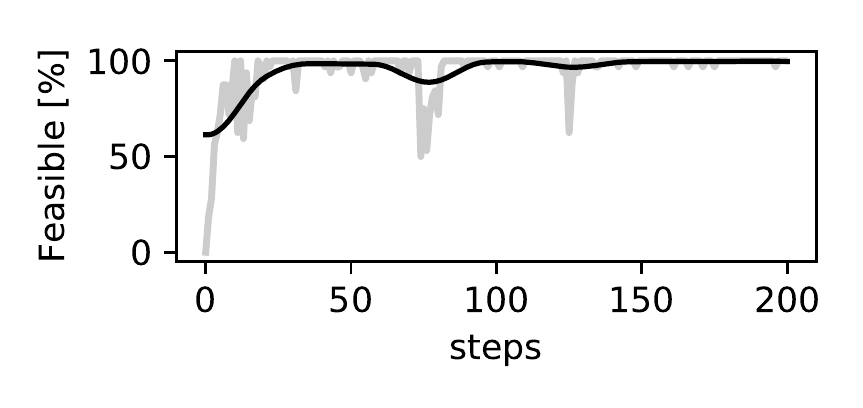}
    \includegraphics[width=0.48\linewidth]{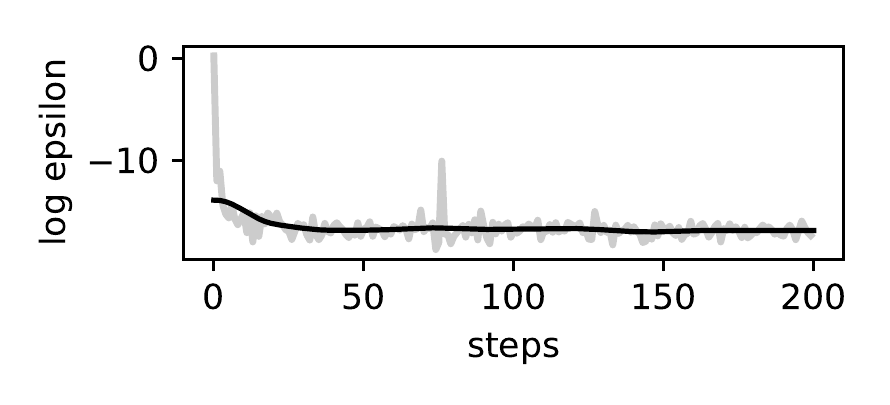}
    \caption[\oursystem: Error and feasibility as a function of learning steps.]{Left: The percentage of feasible solutions as a function of learning steps. Right: The average log mismatch of the power flow equations (\ref{eq:pf_const}) as a function of time.} 
    \label{lopf:violations}
\end{figure*}

\begin{table}[]
\centering
\large
\begin{tabular}{| l | l | l |}
\hline
          & \bf \oursystem & \bf MIPS   \\
          \hline
Feasible [\%] & 99.86  & 60.85 \\
Mean Cost [USD] & 33325.98 & 25817.55\\
Mean time per Instance [s] & 1.2 & 14.4\\
\hline
\end{tabular}
\caption[\oursystem: Performance comparison]{Comparison of \emph{\oursystem} in terms of robustness, optimality and speed on a held-out test set in comparison to MIPS.}
\label{lopf:res_table}
\end{table}

\section{Results}
\label{sec:LOPFresults}
As stated earlier, learning based approaches to control problems are usually not guaranteed to be optimal but can offer advantages in terms of computational time and robustness. The performance of \emph{\oursystem} reinforces these expectations. Figure \ref{lopf:violations} (left) shows the percentage of load flow solutions produced by the system that violate any requirement for feasibility as a function of learning steps. One learning step encompasses 32 load flow problems. For all of the 32 load flow problems 50 candidate binary configurations are drawn from the auxiliary distribution. One can see that the system quickly learns to produce feasible solutions. Initially the system produces feasible solutions to no load flow problems. However, after just 30 steps close to all solutions proposed by the system are feasible.\\
Figure \ref{lopf:violations} (right) shows the $log$-mismatch between the RHS and LHS of the power flow equations (\ref{eq:pf_const}), i.e. $\log \epsilon$. Note that initially, the proposed approach produces generation assignments for which the HELM solver is unable to produce voltage phasors that fulfill the power flow equations but by minimizing the power series coefficients as described in section \ref{sec:enforcing_phys}, the system is quickly nudged into a regime where the proposed solutions fulfill the power flow equations. However, after approximately 75 learning steps, for a short period of time, the system produces generation assignments that, again, do not fulfill the power flow equations. This can most likely be explained by the fact that the system also tries to minimize cost. Thus, by trying to find cheaper generation assignments, the system left the regime in which solutions can be found by HELM but was then steered back into this regime.\\
Table \ref{lopf:res_table} showcases the performance of our proposed algorithm in comparison to the MIPS solver proposed in \cite{zimmerman2011matpower}. In order to deal with non-convex generation limit constraint, the MIPS solver was run with a unit-decommitment heuristic (\emph{runuopf}). When obtaining the results for the MIPS solver, all initializations were unchanged and only demand was varied in the way described earlier. Slightly varying the demand reveals the weakness of traditional solvers: Convergence to a feasible solution cannot be guaranteed. In our experiments, the MIPS solver produced solution which comply with all constraints and fulfill the power flow equations in only about 61\% of all problem instances (failure in 2349 out of 6000 instances). Our proposed solution produces feasible solutions for 99.86\% (failure in 8 out of 6000 cases) of the problem instances.\footnote{When \oursystem~fails, it slightly violates voltage magnitude constraints.}\\
On top of that, our proposed learning based approach is considerably faster than optimization based approaches: Because solutions can be obtained by feeding a demand assignment through the NN
and the forward pass through NNs is usually fast, obtaining the generation assignment proposed by the system is fast. Note that when we report the time per instance for the MIPS solver, we report the mean-time over all load flow problems. However, when the solver fails, it usually fails quickly. If only the time per successful instance was reported, the mean time per instance of the MIPS solver would be close to 30s per instance.\\
However, Table \ref{lopf:res_table} and Figure \ref{lopf:cost} more clearly reveal the main weakness of our proposed learning based approach. Even though solutions can be obtained robustly and fast, the proposed learning system does not find solutions that are optimal in terms of generation cost. On average, the solutions that the approach produces are approximately 29\% more expensive than the solutions found by the MIPS solver. 
Note that the average cost is reported for only those load flow problems for which both approaches yielded feasible solutions.

\section{Conclusion and future work}
\label{sec:LOPFconclusion}
The main contribution of this paper is the introduction of a learning based framework for the problem of ACOPF that respects the full non-linear ACOPF equations. Specifically, we introduce a learning based approach in which a function is tasked to produce feasible and minimal cost generation assignments as a function of the demand. A learning signal for this function is obtained by differentiating through the operators of a load flow solver. Furthermore, we show how convex security constraints and non-convex generation limit constraints can be enforced. The resulting system seems to produce feasible solutions fast. However, these solutions are not necessarily optimal in terms of generation cost.\\
An obvious future research path is to close the optimality gap. At this moment, because of the complexity of the resulting system, it is hard to understand why the solutions are not optimal. But note that the proposed algorithm cannot be optimal by design, because the slack generator cannot be decommitted and there is a natural interpolation between load flow solutions. However, in the opinion of the authors, performance gains in terms of optimality should be possible.\\
Another potentially interesting research question is whether or not the trained auxiliary distribution $Q$ that learns the cost surface as a function of the binary generator configuration allows for conditional sampling. Imagine a scenario where generators have failed. In such a scenario, it is paramount to reconfigure the network in a feasible state fast. If it is possible to sample from $Q$ conditioned that the failed generators are \emph{off}, then the proposed learning based approach could potentially find application in emergency and security sensitive situations. Note that this seemingly easy problem is not trivial because of the FactorNet~\cite{lange2018factornet} structure of the auxiliary distribution.\\

\ifCLASSOPTIONcaptionsoff
  \newpage
\fi

\bibliographystyle{IEEEtran}
\bibliography{lit.bib}

\end{document}